\documentclass{bmvc2k}
\usepackage{multirow}

\title{Detecting Parts for Action Localization}

\addauthor{Nicolas Chesneau}{nicolas.chesneau@inria.fr}{1}
\addauthor{Gr\'egory Rogez}{gregory.rogez@inria.fr}{1}
\addauthor{Karteek Alahari}{karteek.alahari@inria.fr}{1}
\addauthor{Cordelia Schmid}{cordelia.schmid@inria.fr}{1}

\addinstitution{
 Inria$^{*}$
}

\runninghead{Chesneau et al\bmvaOneDot}{Detecting Parts for Action Localization}

\begin{document}
\maketitle

\begin{abstract}
In this paper, we propose a new framework for action localization that tracks
people in videos and extracts full-body human tubes, i.e., spatio-temporal
regions localizing actions, even in the case of occlusions or truncations.
This is achieved by training a novel human part detector that scores visible
parts while regressing full-body bounding boxes. The core of our method is a
convolutional neural network which learns part proposals specific to certain
body parts. These are then combined to detect people robustly in each frame.
Our tracking algorithm connects the image detections temporally to extract
full-body human tubes. We apply our new tube extraction method on the problem
of human action localization, on the popular JHMDB dataset, and a very recent
challenging dataset DALY (Daily Action Localization in YouTube), showing
state-of-the-art results.
\end{abstract}

\section{Introduction}
\label{sec:intro}
Human action recognition in videos is one of the most active fields in computer
vision~\cite{fat,JainCVPR2014,Jhuang:ICCV:2013}. It offers a broad range of
potential applications ranging from surveillance to auto-annotation of movies,
TV footage or sport-videos analysis. Significant progress has been made
recently with the development of deep learning
architectures~\cite{MR2RCNN,Saha2016,WeinzaepfelMS16}. Action localization in
videos comprises recognizing the action as well as locating where it takes
place in the sequence. A popular method for achieving this is to track the
person(s) of interest during the sequence, extract image features in the
resulting ``human tube'' i.e., the sequence of bounding boxes framing a person,
and recognize the action occurring inside the tube. Such a method performs well
when people are fully visible, and when correspondences can be established
between tubes extracted from different videos. This hypothesis does not hold
for most real-world scenarios, e.g., in YouTube videos, where occlusions and
truncations at image boundaries are common and makes action recognition more
challenging. State-of-the-art tracking algorithms~\cite{struck} estimate a
bounding box around the visible parts of a person, resulting in non-homogeneous
tubes that can cover parts of the human body, the full-body or a mix of both in
cases of close-up or moving cameras. For example, in Figure~\ref{fig:general},
a standard human tube extraction method frames the upper-body of the woman
ironing, then the hands and arms, and finally the upper-body again. We posit
that extracting full-body human tubes, even in case of occlusions or
truncations, should help establish better correspondences and extract more
discriminative features, improving action localization performance in complex
scenarios.

The intuition behind our approach is that a bounding box corresponding to the
full human body can often be inferred even if only parts of the person are
visible---scene context and body pose constraints (feasible kinematic
configurations) help estimate where the occluded or truncated body parts are
(see the examples shown in our extracted tubes in Figure~\ref{fig:general}).
To exploit such cues, we propose to train a human part detector that scores the
visible parts but also regresses a full-body bounding box. We present a new
tracking algorithm that simultaneously tracks several parts of a
person-of-interest and combines the corresponding full-body bounding boxes
inferred (regressed) from these parts to reliably localize the full-body. We
demonstrate that our novel tube extraction approach outperforms
state-of-the-art algorithms for action detection and
localization~\cite{WeinzaepfelMS16,weinzaepfelICCV15}.

\vspace{-0.5cm}
\section{Related work}
\vspace{-0.3cm}
\begin{figure*}
\begin{center}
\includegraphics[width=1\linewidth]{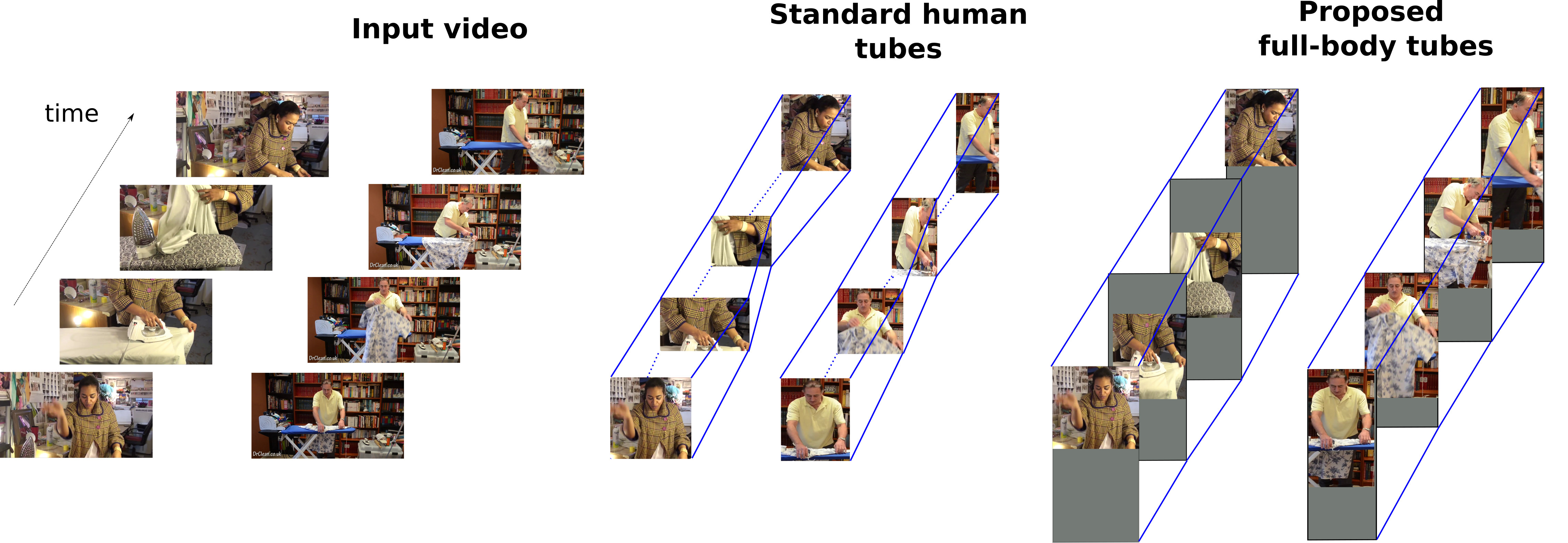}
\end{center}
\vspace{-0.5cm}
\caption{Two example videos from the DALY dataset to illustrate the difference
between our human tube extraction and previous methods (a state-of-the-art
method~\cite{WeinzaepfelMS16} is used for illustration here). Note that our
tubes (shown on the right) take the occluded parts as well as parts beyond the
image boundaries into account.\vspace{-0.5cm}}
\label{fig:general}
\end{figure*}

Initial attempts for temporal and spatio-temporal action localization are based
on a sliding-window scheme and handcrafted
descriptors~\cite{LaptevP07,msr2,yuan2009discriminative,actoms}. Other
approaches, such as~\cite{LanWM11,klaser:inria-00514845}, rely on
figure-centric models, wherein the person performing the action and their
location is detected in some form. In~\cite{LanWM11}, the location of a person
is treated as a latent variable when inferring the action performed, while the
upper body is explicitly detected and tracked in~\cite{klaser:inria-00514845}.
Our approach is also based on human detections but is significantly more robust
to large variations in pose and appearance, due to our learning-based
algorithm. More recently, methods based on action
proposals~\cite{JainCVPR2014,van2015apt,Yu_2015_CVPR,oneata:hal-01021902,marian2015unsupervised}
have been employed to reduce the search complexity and improve the quality of
tracks, referred to as ``tubes''. This paradigm has produced promising results,
but these methods generate thousands of proposals even for a short video
sequence, and are not scalable to large video datasets. Moreover, they do not
take hidden parts and occlusions into account, and are very sensitive to
viewpoint changes. In contrast, our method computes one tube for each person in
the sequence, taking into account body parts that are occluded or truncated by
image boundaries, thereby addressing the problem of amodal
completion~\cite{Kar2015AmodalCA} in the context of action localization.

Recent work has leveraged the success of deep learning for vision tasks in the
context of human action
localization~\cite{fat,weinzaepfelICCV15,MR2RCNN,Saha2016}, by using successful
object detectors, like region proposal-based convolutional neural
networks~\cite{frcnn}. Region-CNNs (R-CNNs) are trained for both appearance and
motion cues in these methods to classify region proposals in individual frames.
Human tubes are then obtained by combining class-specific detections with
either temporal linking based on proximity~\cite{fat}, or with a
tracking-by-detection approach~\cite{weinzaepfelICCV15}. State-of-the-art
methods~\cite{MR2RCNN,Saha2016} rely on an improved version of R-CNN, e.g.,
Faster R-CNN~\cite{frcnn}, trained on appearance and flow. These methods make
extensive use of bounding box annotations in every frame for training the
network. Although this scheme is accurate for short videos, it is not scalable
to long videos with viewpoint changes and close-ups, such as the examples shown
in Figure~\ref{fig:general}. Our method automatically determines the best part
to track and infers the global localization of the person from the part alone.
The merging step of this inference for each part refines the final bounding box
proposed for the frame.

The very recent work in~\cite{WeinzaepfelMS16} is also related to our approach.
It extracts human tubes with a Faster R-CNN based detector. These tubes are
then used to localize and detect actions by combining dense trajectories,
appearance and motion CNN classifiers. However, this method is also limited to
tubes which frame only visible parts of people, and as a result loses spatial
correspondences between frames, thus impacting feature extraction. In contrast,
our method tracks the entire person during the full sequence, making it robust
to partial occlusions, and preserves spatial information for feature extraction
and action classification. We establish new state-of-the-art results on the
challenging DALY dataset proposed in~\cite{WeinzaepfelMS16}, which consists of
31 hours of YouTube videos, with spatial and temporal annotations for 10
everyday human actions.

\vspace{-0.3cm}
\section{Method: From Parts to Tubes }
\begin{figure*}[htb]
\begin{center}
\includegraphics[width=1\linewidth]{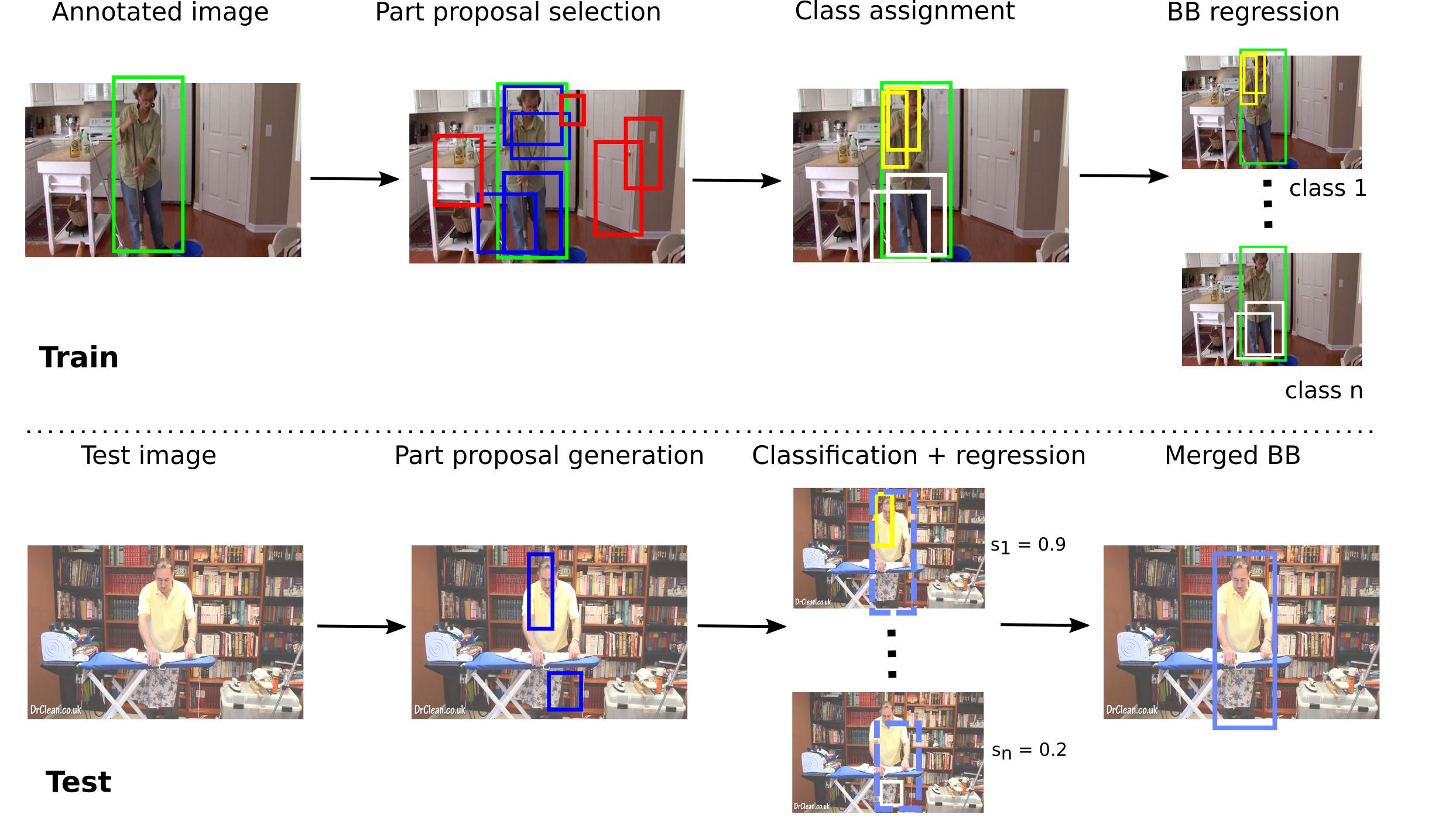}
\end{center}
\vspace{-5mm}
\caption{Overview of our part detector. During training, the detector learns to
select relevant part proposals in the annotated images. These proposals are
assigned a specific  class label and employed to regress the target full-body
bounding box (BB). At test time, the detector generates part proposals, which
are then classified and used to regress full-body bounding boxes. These are
merged to estimate the final bounding box (Merged BB).\vspace{-0.5cm}}
\label{fig:detector}
\end{figure*}

\vspace{-0.3cm}
We propose a new framework for action localization that tracks people in videos
and extract full-body humans tubes, even in case of occlusions or truncations.
This is achieved by training a novel human part detector that scores visible
parts while regressing full-body bounding boxes. Figure~\ref{fig:detector}
shows an overview of our detection architecture that is detailed in
Section~\ref{sec:Dp}. The training phase begins with selection of part
proposals that overlap with groundtruth bounding boxes. These part proposals
are then assigned a particular class label based on their height-width ratio
and location with respect to the bounding box. Finally, a class specific
regressor is trained to infer the full-body bounding box from these parts. At
test time, given an image, we first generate part proposals which are scored,
and then use them to regress full-body bounding boxes, see example in
Figure~\ref{fig:detector}. We then merge these bounding boxes using a new
tracking algorithm, detailed in Section~\ref{sec:BHT}. This algorithm
simultaneously tracks several parts of a person-of-interest and combines the
bounding boxes inferred from these parts to construct a full-body human tube.
Two examples are given in Figure~\ref{fig:general}. Our tube extraction
approach is then employed for action localization as detailed in
Section~\ref{sec:Act}.



\vspace{-0.4cm}
\subsection{Detecting Parts}
\label{sec:Dp}
\vspace{-0.2cm}



Inspired by recent advances of deep learning for computer vision problems, we
propose a CNN-based human part detector which is end-to-end trainable. Given a
database of images annotated with full-body 2D human poses, we first define a
set of human parts from this training data. Then, the learning phase trains our
detector to: (1) generate relevant human part proposals through a region
proposal network (RPN)~\cite{frcnn}, (2) classify and score them, and (3)
regress the corresponding full-body bounding boxes.

\noindent\textbf{Parts definition.} 
In this paper, a human part is defined as a rectangular region covering a small
area of the full-body bounding box. It is represented by its location within
the bounding box (e.g., the ``upper right'' part), an aspect ratio (e.g., a
``square'' or a ``rectangular'' part), and a scale. A set of parts is first
extracted from the training set using a RPN~\cite{frcnn}. The parts must be big
enough to contain relevant information and relatively small to ensure that
multiple different parts will cover the bounding box. To ensure having small
parts, we consider only regions overlapping a groundtruth box with an
intersection over union (IoU) score below a certain threshold. Each part is
then represented by a four-dimensional vector containing its 2D location with
respect to the bounding box center (normalized by bounding box height and
width) and its normalized height and width, expressed as percentages of the
bounding box height and width. The four values are between 0 and 1. A K-means
clustering is finally performed on the set of vectors. The centroids of the
resulting $K$ clusters define our set of part classes. We augment this set of
classes with an additional full-body class.


\noindent\textbf{Parts proposal selection.}
As in~\cite{frcnn}, our architecture uses a RPN to find a set of rectangular
candidate regions. During the learning phase, this set is split into positive
(blue boxes in Figure~\ref{fig:detector}) and negative proposals (red
boxes in Figure~\ref{fig:detector}). We consider a proposal as positive if it
is contained in the groundtruth box and has a fixed number of connected body
keypoints, with two keypoints being connected if they are directly linked on
the human skeleton (e.g., head and shoulders).




\noindent\textbf{Class-specific regression.}
For the class assignment stage, proposals that have a large IoU score with the
groundtruth box are labeled as full-body proposals, while others are assigned
to the closest part class, which is obtained by choosing the corresponding
centroid with the minimum $\ell2$-distance. Note that the groundtruth bounding
box is considered as a positive full-body proposal to include at least one
positive exemplar for each class. The regression step then learns to regress
the 2D image coordinates of the full-body bounding box. This is done
independently for each class to ensure class-specific regression. The goal here
is to localize the rest of the body from a single part. In
the example in Figure~\ref{fig:detector}, two proposal boxes each are assigned
to the classes corresponding to the upper-left area (shown in yellow) and the
legs area (in white). The full-body regression target is shown in green. We
maintain a fixed ratio between part proposals and full-body exemplars in the
training batches.

\noindent\textbf{Test time.}
Our detector first generates relevant part proposals (blue boxes in the lower
part of Figure~\ref{fig:detector}). A full-body bounding box is regressed from
each of these proposals (dashed boxes in the figure). These regressed boxes
corresponding to different part classes can be merged to produce a single
full-body bounding box in a frame with a weighted average, where the weights
are the classification scores. In Figure~\ref{fig:detector}, the yellow box
with a higher score (0.9) has a greater influence than the white box (0.2) on
the final merged bounding box. This produces reasonable detections in several
cases, but we present a more robust approach which leverages all the candidate
boxes for building tubes.

\vspace{-0.4cm}
\subsection{Building full-body tubes}
\label{sec:BHT}
\vspace{-0.2cm}
Given the parts detected, and the corresponding regressed bounding boxes for
the full body, the next task is to build full-body tubes. We perform this by
tracking all the parts detected in each frame, to associate them temporally to
their corresponding parts in other frames, and then use them jointly to
localize the person(s) performing action. To this end, we extend the tracking
algorithm in~\cite{weinzaepfelICCV15}, which is limited to tracking the person
as a whole and can not handle challenging cases where the person is occluded,
as demonstrated in the experimental results (see Section~\ref{sec:results}).

\noindent\textbf{Initialization and tracking the first part.}
We start by detecting body parts in the entire video sequence, as described in
Section~\ref{sec:Dp}. We find the box $b^*$ with the highest score among all
the part classes, and use it to initialize our tracking algorithm. Let this box
be from frame $t$ in the video, and let $B^*$ be the corresponding full-body
(regressed) bounding box. In frame $t+1$, we perform a sliding window search
around the location of the tracked box $b^*$, and select the top scoring box.
This score is a combination of the generic part detector score (given by the
detector in Section~\ref{sec:Dp}) and an instance-level tracker. The
instance-level tracker learns human part appearance in the initialization frame
$t$, with a linear SVM and features from the last fully-connected layer of our
part detector. It is updated every frame with the corresponding chosen box, in
order to handle appearance changes over time. The box $b_{t+1}$ which maximizes
the sum of part detector and instance-level scores, is regressed with our part
detector into a full-body box $B_{t+1}$, and it becomes part of the tube for
frame $t+1$.

\noindent\textbf{Tracking several parts.}
Limiting the tracker to a single body type, which may become occluded in some
of the frames, is prone to missing the person performing an action. To address
this, we detect other parts included into the full-body box $B_{t+1}$ in frame
$t+1$. For example, if the first tracked part is the torso of a person, a
second part could be legs detected by our parts detector. Each detected part is
then tracked independently in following frames, with the method described
above. The regressed full-body boxes of these tracked parts are then combined
to produce one full-body box in each frame, with a weighted average of the sum
of the part detector and instance-level scores.

\vspace{-0.4cm}
\subsection{Action localization}
\label{sec:Act}
\vspace{-0.2cm}
The final step of our approach is to localize actions from the extracted
full-body tubes. We achieve this by representing tubes with features and then
learning an SVM for recognizing actions. We use dense trajectories, RGB and
Flow CNNs as features. A human tube is considered as positive if the temporal
intersection over union (IoU) with the annotated frames is above a certain
threshold. The temporal IoU is defined as the average per-frame IoU. For this
step, our tubes are cropped to frame boundaries to be comparable with
groundtruth annotations. For both RGB and Flow CNN, a R-CNN network is trained
on the respective dataset (i.e., JHMDB and DALY) as follows: region proposals
in a frame whose IoU with our estimated bounding box is above a threshold are
labeled as positives for the class of the tube. Action classifiers are learned
for each of the features independently and then combined using a late-fusion
strategy. These steps are described in detail in Section~\ref{sec:implement}.

\begin{figure}[t]
\begin{center}
\begin{tabular}{c c c c}
\includegraphics[width=0.22\linewidth]{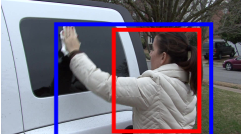}&
\includegraphics[width=0.22\linewidth]{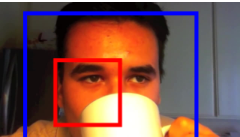}&
\includegraphics[width=0.22\linewidth]{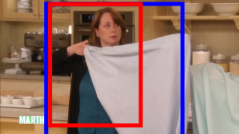}&
\includegraphics[width=0.22\linewidth]{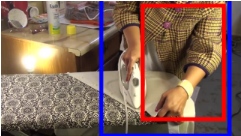}
\end{tabular}
\end{center}
\vspace{-0.3cm}
\caption{Qualitative results. The visible part of our full-body tubes is shown
in blue. For comparison, the tubes of the state-of-the-art
method~\cite{WeinzaepfelMS16} are in red. Here, we show a sample frame from
four example videos corresponding to \textit{Cleaning windows},
\textit{Drinking}, \textit{Folding Textile}, and \textit{Ironing} events of
DALY.\vspace{-0.4cm}}
\label{fig:illustration}
\end{figure}

\vspace{-0.4cm}
\section{Experiments}
\label{sec:xp}
\vspace{-0.2cm}

\subsection{Datasets}
\vspace{-0.2cm}
\noindent\textbf{Test data.} We evaluate our method on  two action recognition
datasets, namely JHMDB~\cite{Jhuang:ICCV:2013} and DALY~\cite{WeinzaepfelMS16}.
JHMDB is a standard action recognition database used
in~\cite{Saha2016,MR2RCNN,weinzaepfelICCV15,fat,WeinzaepfelMS16}. It is a
subset of the larger HMDB51 dataset collected from digitized movies and YouTube
videos. It contains 928 videos covering 21 action classes. DALY is a more
challenging large-scale action localization dataset consisting of 31 hours of
YouTube videos (3.3M frames, 3.6k instances) with spatial and temporal
annotations for 10 everyday human actions. Each video lasts between 1 and 20
minutes with an average duration of 3min 45s. We also use the LSP
dataset~\cite{Johnson10} for analyzing our full-body box generation method. LSP
contains 2000 pose annotated images of mostly sports people gathered from
Flickr.

\noindent\textbf{Training data.}
For training our human part detector, we use the MPII human pose
dataset~\cite{andriluka14cvpr}. It covers around 400 actions and contains a
wide range of camera viewpoints. The training set consists of around 17k
images, with each scene containing at least one person, often occluded or
truncated at frame boundaries. As in~\cite{WeinzaepfelMS16}, we compute a
groundtruth bounding box for each person by taking the box containing all
annotated body keypoints with a fixed additional margin of 20 pixels. 
To obtain
full-body bounding boxes, in cases of occlusions and truncations (where only
visible key points are annotated), we employ a nearest neighbor search on the annotated keypoints to
complete missing annotations and recover complete full-body 2D poses.
As done in \cite{rogez:hal-01505085}, we generate a large set (8M) of human 2D poses by projecting 3D poses from the CMU Motion Capture dataset on multiple random camera views.
Then, for each incomplete 2D pose in the MPII training set, a search is
performed on the annotated 2D joints to estimate the closest match, i.e.,
full-body 2D pose, that is later employed to estimate a full-body bounding box.

\begin{figure*}
\begin{center}
\includegraphics[width=0.9\linewidth]{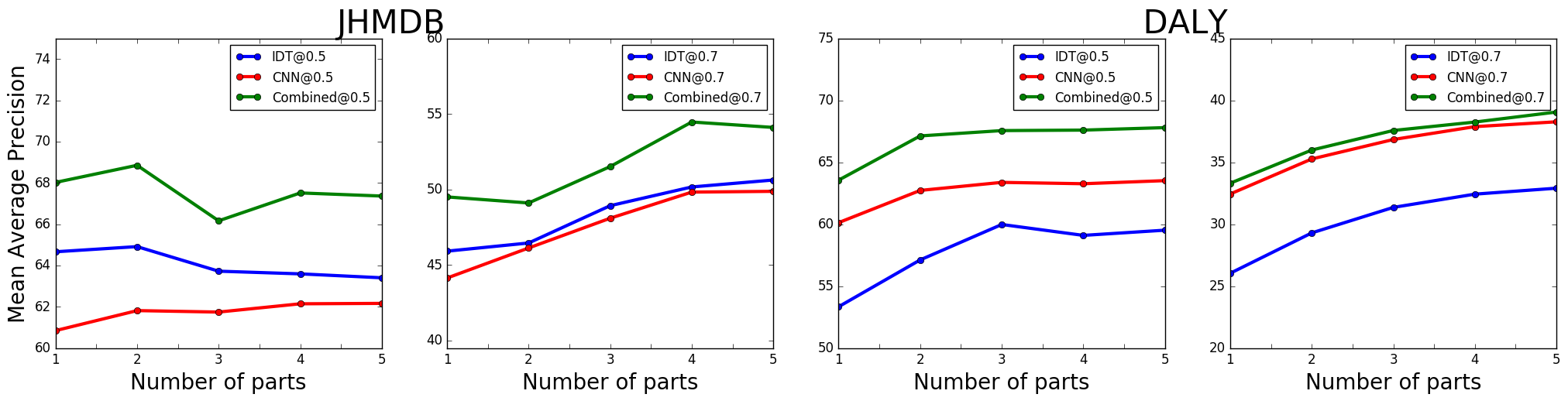}
\end{center}
\vspace{-0.5cm}
\caption{mAP@0.5 and mAP@0.7 results on JHMDB and DALY datasets with respect to
the number of parts used for building human tubes.\vspace{-0.4cm}}
\label{fig:parts_xp}
\end{figure*}

\begin{figure*}
\begin{center}
\includegraphics[width=0.9\linewidth]{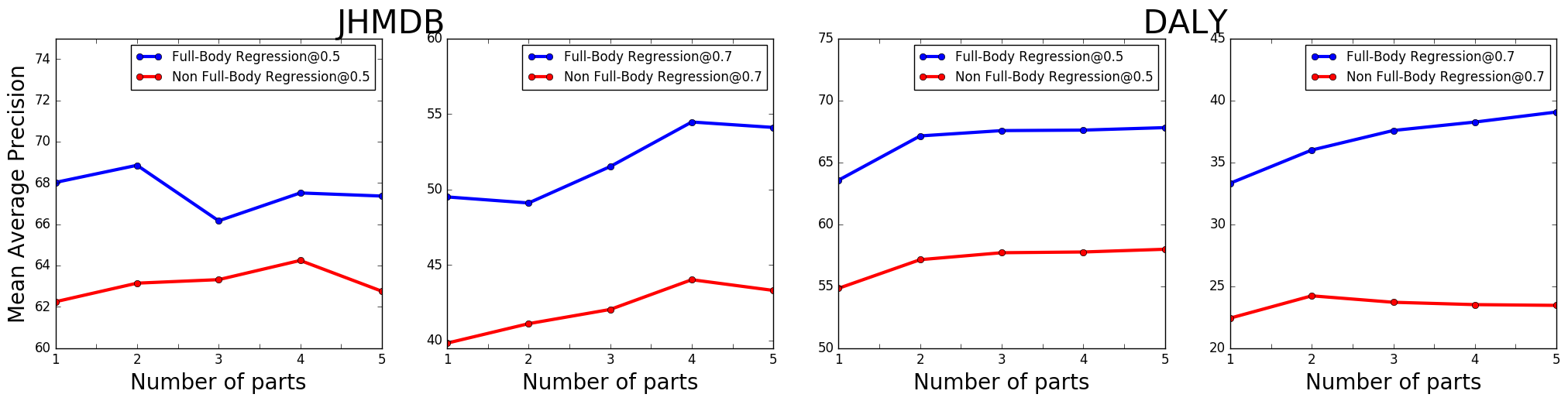}
\end{center}
\vspace{-0.5cm}
\caption{Comparison of our method (trained with full-body boxes) and a variant
that only uses visible regions. mAP@0.5 and mAP@0.7 results on JHMDB and DALY
datasets with respect to the number of parts used for tube generation are
shown.\vspace{-0.4cm}}
\label{fig:comp_NB}
\end{figure*}

\vspace{-0.5cm}
\subsection{Implementation details}
\label{sec:implement}
\vspace{-0.2cm}
The implementation of our part detector is based on Faster-RCNN~\cite{frcnn}
with VGG16  layers~\cite{DBLP:journals/corr/SimonyanZ14a}. The number of
classes is set to 21: 20  human parts and the full body class. 



\noindent\textbf{Part detector.}
Keypoints are annotated in the MPII dataset we use to train the
detector. The connections between joints are defined following the standard
human skeleton (e.g., head connected to shoulders, shoulders to elbows). For
the part proposal selection stage, we consider a proposal as positive if it
overlaps the groundtruth box and contains exactly three connected body
keypoints. We tested with different numbers of keypoints and found that three
was an optimum number to maximize human detection rate on DALY (with
recall@0.5). For the class assignment stage, proposals with IoU more than
$0.55$ are labeled as full-body proposals, while those with IoU between $0.1$
and $0.55$ are assigned to the class of the closest part.


\noindent\textbf{Optimization.}
Initialization of the network is done with ImageNet pretrained weights. The
number of iterations is set to 180K, the learning rate to 0.001, the momentum
to 0.9, the gamma parameter to 0.1, i.e., the learning rate is divided by 10 at
every learning step (@ 100K, 150K, 170K iterations). We use batches of 128
proposals (32 positive and 96 background) and constrain each bach to have 10
times more part proposals than full-body exemplars.

\noindent\textbf{Human tubes.} 
For the DALY dataset, tubes are computed using every fifth frame for
computational reasons. During the tracking procedure, a box is removed if its
combined score (defined in Section~\ref{sec:BHT}) is less than $1$. A new part
is added to the tube if it has a detector score of $ 0.25$.

\noindent\textbf{Action localization.}
The dimensions of the four descriptors (HOG, HOF, MBHx and MBHy) are reduced by
a factor of 2 using PCA and a codebook of 256 Gaussians. Appearance and motion
CNNs are based on R-CNN architecture proposed in~\cite{girshick2014rcnn}.
Five annotated frames per sequence are used as for computing the temporal IoU with our tubes. A human tube is considered as
positive if the temporal intersection over union (IoU) with the five annotated
frames is above 0.5. During training, the region proposals whose IoU with our estimated bounding box
is above 0.5 are labeled as positives for the class of the tube. One linear SVM
classifier each is learned independently for the three feature representations
(dense trajectory, appearance and motion CNNs). . During test time, scores are scaled between 0 and 1 using
a sigmoid, and the global score of a tube is the sum of the three SVM scores.

\begin{table}
\begin{center}
\resizebox{\linewidth}{!}{
\begin{tabular}{|l|l|c|c|c|c|}
\hline
\multirow{2}{*}{Features}& \multirow{2}{*}{Method}&  \multicolumn{2}{c|}{DALY}  & \multicolumn{2}{c|}{JHMDB}  \\\cline{3-6}
                                    & & meanAP@0.5 & meanAP@0.7  & meanAP@0.5 & meanAP@0.7   \\
\hline
\multirow{2}{*}{Dense Trajectories} & Ours                   & 58.97 & 31.35 & 64.91 & 46.45 \\
                                    & \cite{WeinzaepfelMS16} & 53.21 & 21.57 & 60.11 & 41.39 \\
\hline
\multirow{2}{*}{Appearance \& motion CNNs} & Ours                    & 63.51 & 38.21 & 61.81 & 46.12 \\
                                           & \cite{WeinzaepfelMS16}  & 61.12 & 28.37 & 64.08 & 49.22 \\
                                           & \cite{fat}              &   -   &   -   & 53.30 &   -   \\
                                           & \cite{weinzaepfelICCV15}&   -   &   -   & 60.70 &   -   \\
                                           & \cite{MR2RCNN}          &   -   &   -   & 73.10 &   -   \\
                                           & \cite{Saha2016}         &   -   &   -   & 71.50 &   -   \\
\hline
\multirow{2}{*}{Combination} & Ours                   & 67.79 & 39.05 & 68.85 & 49.10 \\
                             & \cite{WeinzaepfelMS16} & 64.56 & 29.31 & 65.80 & 49.54 \\
\hline
\end{tabular}
}
\end{center}
\vspace{-0.3cm}
\caption{Comparison to the state of the art with mAP@0.5 and mAP@0.7 measures
on DALY and JHMDB datasets. We report results for the fully-supervised variant
of~\cite{WeinzaepfelMS16}.\vspace{-0.3cm}}
\label{tab:res}
\end{table}

\vspace{-0.5cm}
\subsection{Results}
\label{sec:results}
In Table \ref{tab:res}, our method shows an improvement
over~\cite{WeinzaepfelMS16} on both DALY and JHMDB, of 3.21\% and 3.05\%
respectively for mAP@0.5. A larger gain is obtained with mAP@0.7 on DALY
(9.74\% for ``Combination''), showing that our method is more accurate for
action localization in videos. All the results of~\cite{WeinzaepfelMS16} were
obtained directly from the authors. For AP@0.5, per event results emphasize the
role of detecting and tracking multiple parts (see
Table~\ref{tab:per-eventsres}). Compared to~\cite{WeinzaepfelMS16}, we
significantly improve the performance for actions such as \textit{Applying make
up on lips} with 81.91\% (vs 68.18\% for~\cite{WeinzaepfelMS16}),
\textit{Brushing teeth} with 68.64\% (vs 57.61\%).  Videos of these actions are
often close-up views, where the body is not fully visible during all or part of
the sequence. This makes computing feature correspondences between frames more
difficult for methods such as~\cite{WeinzaepfelMS16} which do not estimate the
full-body bounding box. The difference between the two methods is even more
important for AP@0.7: 49.27\% (vs 2.62\%) for \textit{Applying make up on
lips}, 28.62\% (vs 20.58\%) for \textit{Brushing teeth}. Our human tubes
estimate the position of the full body and infer the location of non-visible
parts. This provides a canonical region of support for aggregating local
descriptors which belongs to the same parts of the body. Although the method
in~\cite{MR2RCNN} shows better results on JHMDB, our method has the advantage
of being scalable to larger datasets and longer videos. It can also be applied
in a weakly-supervised way.

Figures~\ref{fig:detector} and~\ref{fig:illustration} show a selection of
qualitative results. Although Figure~\ref{fig:detector} shows standing persons,
people seated are also well-detected. For example, for the \textit{Playing
Harmonica} event in DALY, which contains videos of people sitting (34 examples)
and standing (16), we observe a significant improvement: over 1.8\% and 31\%
for AP@0.5 and AP@0.7 respectively. Figure~\ref{fig:illustration} compares our
tubes with those extracted by~\cite{WeinzaepfelMS16}, showing that our method
better handles close-up views and occlusions.

\noindent\textbf{Influence of part trackers.} Figure~\ref{fig:parts_xp} shows
the mean average precision of our method when varying the number of parts being
tracked when building our tubes. The gain of adding parts is particularly
significant with AP@0.7 for JHMDB. For AP@0.5, two-part tracking gives the best
results because videos are short and viewpoint changes are limited. For AP@0.7,
tracking a maximum of four parts improves average precision significantly. On
average, we obtain an mAP of 54.46\%, with an improvement of 4.92\%
over~\cite{WeinzaepfelMS16}. The results for a few specific actions highlight
the effectiveness of our tracking. For example, the \textit{wave} action has an
average precision of 32.49\% with 1-part tracking, and 49.92\% when tracking 4
parts. Videos of this action contain two different points of view: a
``full-body'' point of view, and a ``torso'' point of view, making the use of
full-body tubes relevant and effective. A similar observation can be made with
the \textit{climb stairs} action (``full body'' and ``legs'' points of view),
with a gain of 8.27\% (50.80\% with single-part tracking, 59.07\% with 4
parts), and also for the \textit{throw} action, with a significant number of
upper-body and head videos (18.74 \% with single-part tracking, 45.90\% with
4). On the contrary, the \textit{walk} action shows better results with a
single-part tracker (67.56\%, compared to 64.59\%) because the body of the
person walking is fully visible in all the videos, and using the full-body
class suffices. On the DALY dataset, a five-part tracker gives the best
results. This is partly due to DALY being a much more challenging dataset than
JHMDB.

\begin{table}
\begin{center}
\resizebox{0.6\textwidth}{!}{
\begin{tabular}{|l|l|c|c|c|c|}
\hline
\multirow{2}{*}{Classes}&  \multicolumn{2}{c|}{\cite{WeinzaepfelMS16}}  & \multicolumn{2}{c|}{Ours}  \\\cline{2-5}

& AP@0.5 & AP@0.7  & AP@0.5 & AP@0.7   \\
\hline
 ApplyingMakeUpOnLips & 68.18 & 2.62 &81.91 &49.27\\
 BrushingTeeth& 57.61 & 20.58 &68.64 &28.62 \\
 CleaningFloor& 88.54 & 72.56 &86.13 &66.85 \\
 CleaningWindows& 77.37 & 35.25 &78.13 &53.05 \\
 Drinking& 44.10& 13.99 &36.77 &24.86 \\
 FoldingTextile& 58.90 & 35.35 &60.77&15.94\\
 Ironing& 78.28 & 39.38 &82.52&29.68 \\
 Phoning& 52.06 & 25.05 &63.41&34.19\\
 PlayingHarmonica& 68.36 & 26.93 &70.18&58.12 \\
 TakingPhotosOrVideos& 52.19 & 21.36 &49.42 &29.95\\
 Mean& 64.56 & 29.31 &67.79 &39.05 \\       
\hline
\end{tabular}
}
\end{center}
\vspace{-0.2cm}
\caption{Per-event results on the DALY dataset of our method
and~\cite{WeinzaepfelMS16}. The results of our method correspond to the one
using five parts for human tracking (see
Section~\ref{sec:BHT}).\vspace{-0.4cm}}
\label{tab:per-eventsres}
\end{table}

\noindent\textbf{Influence of fully-body tubes.}
Figure~\ref{fig:comp_NB} compares the performance of part
detectors that regress to full-body (including occluded or truncated regions)
vs those that regress only to visible regions (i.e., non full-body). Building
non full-body tubes decreases the performance, for example, from 68.85 to 63.14
on JHMDB, from 67.79 to 57.97 on DALY for AP@0.5.  It confirms our idea that
building full-body tubes instead of the standard ones are well-adapted for
action localization and classification, and can: (1) establish better feature
correspondences, and (2) better exploit techniques such as spatial pyramid
matching (SPM) for recognition tasks. Additional experiments show that SPM is
more effective with dense tracks when considering our full-body tubes (+3\%
mAP) vs cropped and mis-aligned tubes from~\cite{WeinzaepfelMS16} (+1\%). In
essence, such an ``amodal completion'' defines a better reference frame for
features (spatio-temporal grid is more adapted as person-centric), and results
in better performance in the context of action localization.

\begin{figure}[t]
\begin{center}
\resizebox{\linewidth}{!}{
\begin{tabular}{@{}c @{}c}
\includegraphics[width=0.75\linewidth]{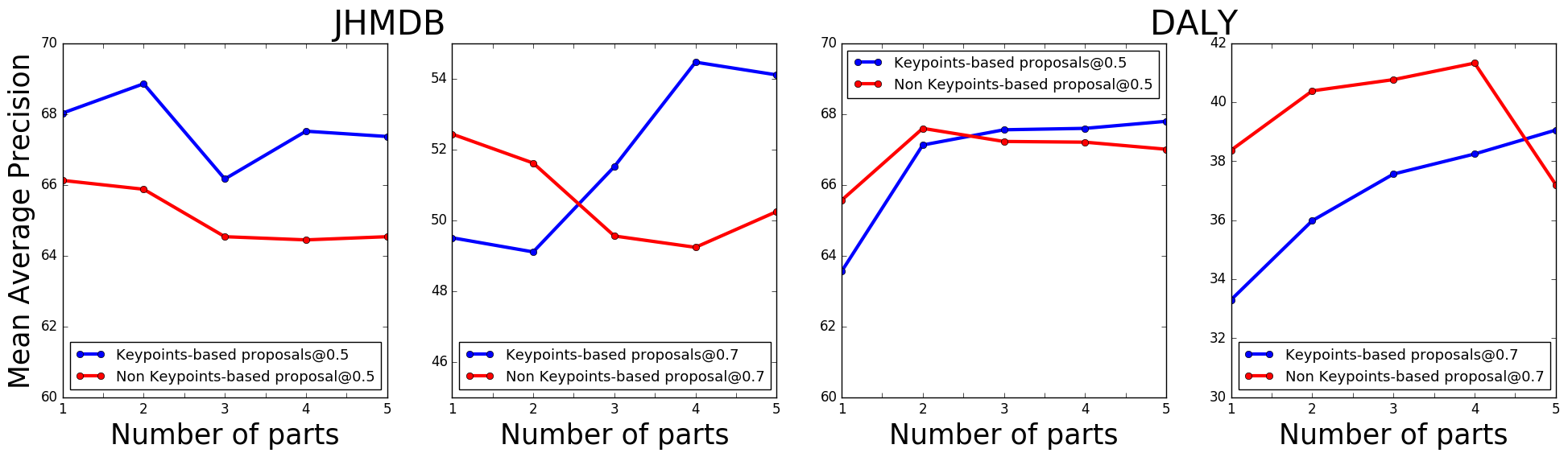}&
\includegraphics[width=0.25\linewidth]{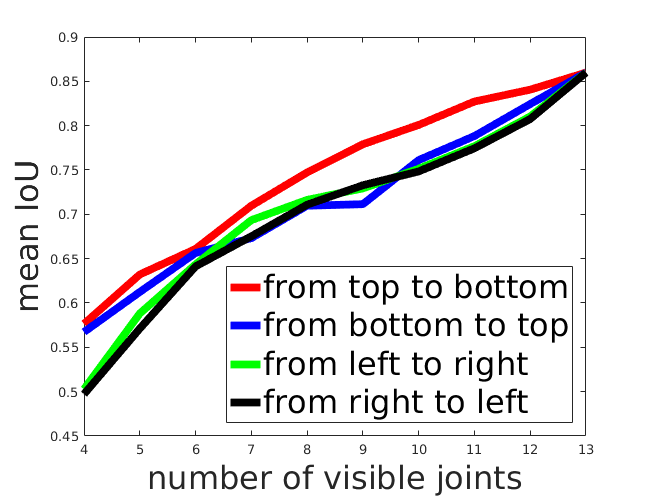} \\
(a)&(b)
\end{tabular}
}
\end{center}
\vspace{-0.3cm}
\caption{(a) Influence of keypoint based proposal generation. (b) Mean IoU of
our full-body box generation method and groundtruth boxes on the LSP dataset,
with respect to the number of visible keypoints.\vspace{-0.3cm}}
\label{fig:cam_read_add}
\end{figure}

\noindent\textbf{Influence of keypoints.} The results in Figure~\ref{fig:cam_read_add}(a) highlight the
importance of keypoint based proposal generation. We compare our full method,
which uses keypoints for selecting parts proposals (refer Section~\ref{sec:Dp})
with a variant that considers a proposal as positive if its overlap with ground
truth is in the range of 0.2 and 0.6, i.e., without using keypoints. The
performance of this no-keypoint variant is lower than our full method: 65.87 vs
68.85 on JHMDB, and 67.22 vs 67.79 on DALY.

\noindent\textbf{Analysis of full-body box generation.} We
simulated partially-occluded human poses on the LSP dataset for this analysis.
Given a full pose, we successively remove the lowest keypoint in the human
pose/skeleton, then the two lowest keypoints, and so on. We then estimate the
full-body box with our method for each of these simulated incomplete poses,
i.e, the box which frames the full estimated pose. The effectiveness of this
estimation is measured by comparing it with the groundtruth box. We do this
experiment by removing successively the highest, the left-most, and the
right-most keypoints. Results are shown in Figure~\ref{fig:cam_read_add}(b).
Full-body boxes estimated with 9 out of the 13 keypoints (which corresponds to
missing legs for the ``lowest" experiment, and missing head and shoulders for
the ``highest" experiment) gives a mean IoU over 0.7 with the groundtruth in
all the four cases. With only 4 out of 13 keypoints, the IoU remains relatively
high (between 0.5 and 0.6).

\noindent\textbf{Analysis of annotations.} Although our method shows
state-of-the-art results on action localization, it suffers from an annotation
bias. Our human tubes frame the full body, including hidden parts. For the {\it
Ironing} event in DALY, legs are frequently hidden by ironing boards (see
Figure~\ref{fig:general} with two examples of DALY videos where the ironing
event occurs), and the annotations are focused on visible parts of the body,
i.e., the torso and arms. Consequently, the IoUs between our tubes and
annotations suffer from the fact that they do not cover the same parts of the
actors. To estimate the impact of this annotation bias, we re-annotated the
groundtruth bounding boxes in all {\it Ironing} sequences from DALY taking into
account the hidden body parts. We then computed the AP@0.5 and AP@0.7 with
dense trajectories and CNNs. With these new annotations, we obtain an average
precision of 85.2\% (45.55\% for AP@0.7), whereas average precision with
original annotations is 82.52\% (29.68\% for AP@0.7). The method
in~\cite{WeinzaepfelMS16} obtains an average precision of 78.28\% (39.38\% for
AP@0.7) with the original
annotations. The experiment shows that the classical way of annotating humans
in computer vision datasets, i.e., annotating only visible parts, is not ideal
to correctly evaluate our full-body tube extractor. However, our results show
significant improvements in action localization in a semi-supervised way.
Although we train our part detector for people detection, it can be extended to
all objects. The main point is to have a training dataset with annotations for
the full object, i.e., taking into account hidden parts. Our part detector can
also be used with tracking by detection algorithms~\cite{Saha2016}.

\vspace{-0.4cm}
\section{Conclusion}
\label{sec:conc}
\vspace{-0.3cm}
We proposed a novel full-body tube extraction method based on a new body part
detector. These detectors are specific to body parts, but regress to full-body
bounding boxes, thus localizing the person(s) in a video. Our tube extraction
method tracks several human parts through time, handling occlusions, view point
changes, and localizes the full body in any of these challenging scenarios. We
showed that using our full-body tubes significantly improves action
localization compared to methods focusing on tubes built from visible parts
only, with state-of-the art results on the new challenging DALY dataset.

{\small\paragraph{Acknowledgments.} This work was supported in part by the ERC advanced grant ALLEGRO, the Indo-French project EVEREST funded by CEFIPRA and an Amazon research award. We gratefully acknowledge the support of NVIDIA with the donation of GPUs used for this research.}
\bibliography{egbib}

\end{document}